\documentclass[runningheads]{llncs}

\usepackage{eccv}

\usepackage{eccvabbrv}

\usepackage{graphicx}
\usepackage{booktabs}

\usepackage[accsupp]{axessibility}  

\usepackage{hyperref}

\usepackage{orcidlink}
\usepackage{graphicx}
\usepackage{amsmath}
\usepackage{amssymb}
\usepackage{booktabs}
\usepackage{mathtools}
\usepackage{multirow}
\usepackage{caption}
\usepackage{pifont}
\usepackage{url}
\newcommand{\cmark}{\ding{51}}%
\newcommand{\xmark}{\ding{55}}%
\newcommand{\tit}[1]{\smallbreak\noindent\textbf{#1.}}

\begin{document}

\title{WildVidFit: Video Virtual Try-On in the Wild via Image-Based Controlled Diffusion Models}

\titlerunning{WildVidFit: Video Virtual Try-On in the Wild}

\author{Zijian He\inst{1}\orcidlink{0009-0003-1845-2195} \and
Peixin Chen\inst{1}\orcidlink{0009-0002-8641-1048} \and
Guangrun Wang\inst{1}\orcidlink{0000-0001-7760-1339} \and 
Guanbin Li\inst{1,2}\thanks{Corresponding Author}\orcidlink{0000-0002-4805-0926} \and 
Philip H.S. Torr\inst{3}\orcidlink{0009-0006-0259-5732} \and
Liang Lin\inst{1,2}\orcidlink{0000-0003-2248-3755}}

\authorrunning{Z. He et al.}

\institute{Sun Yat-sen University\and
Peng Cheng Laboratory\and
University of Oxford\\
\email{\{hezj39,chenpx28\}@mail2.sysu.edu.cn}, wanggrun@gmail.com, liguanbin@mail.sysu.edu.cn, philip.torr@eng.ox.ac.uk, linliang@ieee.org}

\maketitle
\begin{center}
    \captionsetup{type=figure}
    \includegraphics[width=0.9\textwidth]{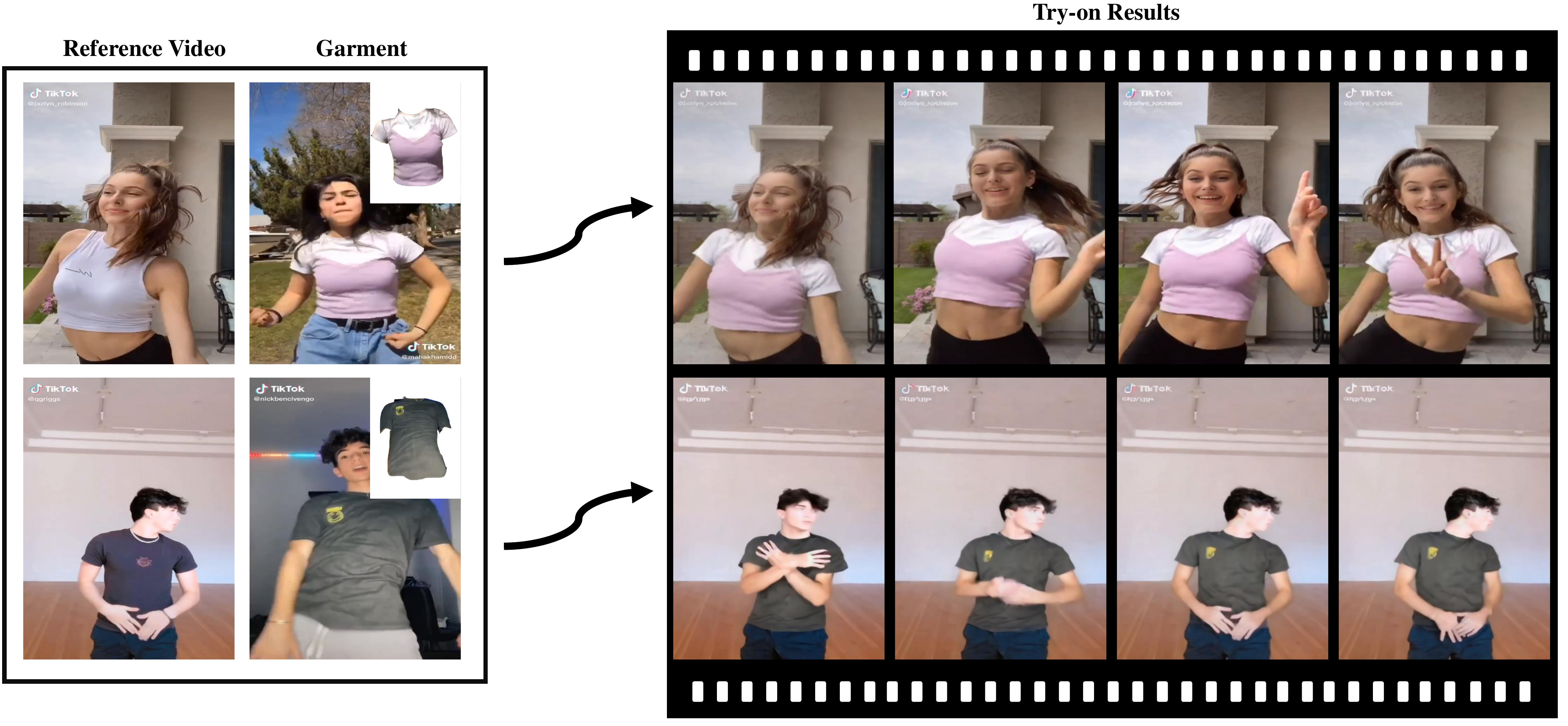}
    \captionof{figure}{Examples of our virtual try-on results on real-life TikTok videos.} 
\end{center}

\begin{abstract}
    Video virtual try-on aims to generate realistic sequences that maintain garment identity and adapt to a person's pose and body shape in source videos. Traditional image-based methods, relying on warping and blending, struggle with complex human movements and occlusions, limiting their effectiveness in video try-on applications. Moreover, video-based models require extensive, high-quality data and substantial computational resources. To tackle these issues, we reconceptualize video try-on as a process of generating videos conditioned on garment descriptions and human motion. Our solution, WildVidFit, employs image-based controlled diffusion models for a streamlined, one-stage approach. This model, conditioned on specific garments and individuals, is trained on still images rather than videos. It leverages diffusion guidance from pre-trained models including a video masked autoencoder for segment smoothness improvement and a self-supervised model for feature alignment of adjacent frame in the latent space. This integration markedly boosts the model's ability to maintain temporal coherence, enabling more effective video try-on within an image-based framework. Our experiments on the VITON-HD and DressCode datasets, along with tests on the VVT and TikTok datasets, demonstrate WildVidFit's capability to generate fluid and coherent videos. The project page website is at \url{wildvidfit-project.github.io}. 
\end{abstract}

\section{Introduction}
\label{sec:intro}


In video virtual try-on, the objective is to generate seamless videos that preserve the appearance of a specific garment while accurately adapting to the pose and body shape of the individual in the source video. This domain has garnered significant attention due to its potential applications in e-commerce and the burgeoning short-form video sector.

Recent advancements in video virtual try-on have evolved from initial two-stage image-based approaches, involving flow-based warping and blending, to incorporating an additional temporal module for ensuring frame consistency. Notably, FW-GAN~\cite{dong2019fw} introduced an optical flow-guided fusion module, utilizing past frame warping results for current frame prediction. MV-TON~\cite{zhong2021mv} employed garment-to-person flow estimation for each frame, coupled with a memory module for refining frames using space-time information. ClothFormer~\cite{jiang2022clothformer} achieved realistic, spatio-temporally consistent results with its anti-occlusion warping, appearance-flow tracking, and dual-stream transformer. However, these methods face challenges in ``in-the-wild'' video applications due to two primary obstacles. Firstly, the collection of robust video data is costly, and developing a temporal module requires extensive, high-quality videos and computational resources. These methods, trained on specific datasets~\cite{dong2019fw,jiang2022clothformer}, have limited generalization ability. Secondly, limb occlusions and significant garment deformation, more prevalent in videos than still images, lead to misalignment issues in current methodologies.

Addressing the video virtual try-on challenge hinges on generating images that adhere to given conditions, such as garment descriptions and human motion sequences. Particularly for videos captured in uncontrolled environments, a method must robustly handle intricate motions and limb occlusions. An image-based approach, leveraging extensive image foundation model knowledge and abundant image data, is particularly beneficial. This leads to decomposing the video try-on task into two subtasks: developing a fine-grained image try-on model for complex movements and occlusions, and extending it to video while maintaining frame coherence. 

In response, we introduce WildVidFit, a novel, video training-free virtual try-on framework. WildVidFit utilizes image-based controlled diffusion models for realistic video try-on results. It bypasses explicit warping limitations in occlusion handling with a detail-focused, one-stage image try-on network, synthesizing outputs based on unified representations of garments and individuals. It incorporates implicit warping inspired by TryOnDiffusion~\cite{zhu2023tryondiffusion} for naturalistic outcomes and features a diffusion guidance module. This module enhances the temporal consistency of videos by improving segment smoothness with a pre-trained video masked autoencoder and aligning features of adjacent frames in the latent space through a self-supervised model. Crucially, WildVidFit streamlines the process by using editing and content consistency cues from pre-trained models, eliminating the need for additional fine-tuning or new temporal modules. Our contributions can be summarized as follows:
\begin{itemize}
    \item We present WildVidFit, a video training-free virtual try-on framework capable of handling complex limb occlusions and actions in wild videos with a straightforward process.    
    \item We introduce a diffusion guidance module to enhance temporal consistency, employing pre-trained video models and image self-supervised models to establish frame feature correspondence in the latent space. 
    \item 
    Our experiments on the TikTok dataset demonstrate WildVidFit's effectiveness in dynamic, real-world scenarios, highlighting its practicality and versatility.
\end{itemize}




\section{Related Work}

\noindent\textbf{Image Virtual Try On.}
Given a pair of images (reference person, target garment), image virtual try-on methods aim to generate the appearance of the reference person wearing the target garment. Most of these methods~\cite{han2018viton,wang2018toward,yu2019vtnfp,issenhuth2020not,yang2020towards,choi2021viton,ge2021parser,dong2022dressing,He_2022_CVPR,Yang_2022_CVPR,lee2022hrviton,bai2022single,men2020controllable,zhang2021pise,ren2022neural} decompose the try-on task into two generation stages,  i.e., warping and blending. The pioneering work, VITON~\cite{han2018viton}, introduced a coarse-to-fine pipeline that was guided by the thin-plate-spline (TPS) warping of the target garment. ClothFlow~\cite{han2019clothflow} advanced the warping process by directly estimating the flow field using a neural network instead of the TPS. VITON-HD~\cite{choi2021viton} released a high-resolution virtual try-on dataset and increased the resolution of generated images from $256\times 192$ to $1024\times 768$ with an alignment-aware generator. GP-VITON~\cite{xie2023gp} developed an innovative Local-Flow-Global-Parsing warping module to preserve the semantic information of different parts of the garment. Moreover, \cite{huang2022towards} integrated geometric priors of 3D human bodies, enabling a more nuanced handling of pose and viewpoint variations. Although these methods have made significant progress, explicit warping still struggles to cope with complex poses and occlusions due to pixel misalignment.

Recently, diffusion models~\cite{sohl2015deep,song2019generative,ho2020denoising} have risen to prominence as the leading family of generative models. As a result, there is a growing interest in leveraging diffusion models as an alternative to GANs to achieve more realistic outcomes. LaDI-VTON~\cite{morelli2023ladi} incorporated the latent diffusion model~\cite{rombach2022high} into the blending stage of virtual try-on and introduced a textual inversion module to enhance the texture on garments. DCI-VTON~\cite{gou2023taming} proposed an exemplar-based inpainting approach that leveraged a warping module to guide the diffusion model's generation. Both methods follow the previous two-stage approach. TryOnDiffusion~\cite{zhu2023tryondiffusion} presented a diffusion-based architecture, enabling the preservation of garment details and the ability to warp the garment to accommodate significant pose and body changes within a single network. However, the network design of two parallel UNets followed by a super-resolution module will bring huge computation cost when extending TryOnDiffusion to video synthesis.

\tit{Video Virtual Try On}
Researches extended the two-stage approach used in image virtual try-on to video applications by integrating a specially designed temporal module. FW-GAN~\cite{dong2019fw} successfully applied a video generation framework to the task of virtual try-on by incorporating relevant factors like warped garments and human postures. MVTON~\cite{zhong2021mv} introduced a try-on module for garment warping using pose alignment and regional pixel displace, and a memory refinement module that embedded prior generated frames into a latent space, serving as external memory for subsequent frame generation. ClothFormer~\cite{jiang2022clothformer}, on the other hand, refined flow predictions using inter-frame information and employed a Dual-Stream Transformer to produce the video try-on result from warping results of multiple frames. Despite great advancements in walking scenarios, these methods face challenges when applied to wild videos featuring complex human movements. This is attributed to the high cost of labeled video data and the inherent limitations of explicit warping.
\section{Method}
\begin{figure} [ht]
    \centering
    \includegraphics[width=1.0\linewidth]{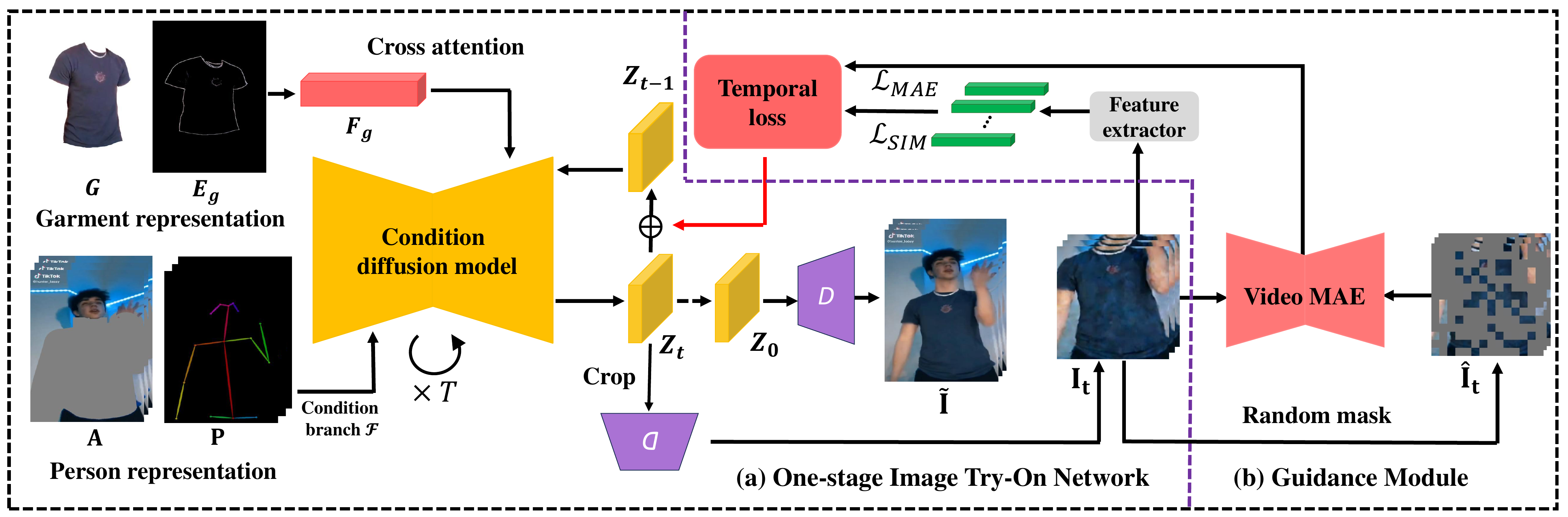} 
    \caption{\textbf{Overview of our WildVidFit framework.} Our method contains two modules, i.e., a one-stage image try-on network and a guidance module. 
    In timestep $t$, we crop the garment area and decode the latent $Z_t$ into sequence $\mathbf{I_t}$. The similarity loss $L_{SIM}$ is calculated between adjacent frames $I^{j+1}_t$ and $I^j_t$ using spherical distance. Additionally, we randomly mask the sequence $\mathbf{I_t}$ into $\hat{\mathbf{I}}_t$, which is then inputted into VideoMAE for reconstruction. $L_{MAE}$ represents the distance between the sequences $\mathbf{I_t}$ and $\hat{\mathbf{I}}_t$. We assume that a lower reconstruction loss will result in a smoother sequence. $L_{SIM}$ and $L_{MAE}$ together constitute the temporal loss, which controls the sampling process from $Z_t$ to $Z_{t-1}$.}
    \label{fig:guidance}
\end{figure}

Fig.~\ref{fig:guidance} provides an overview of our proposed WildVidFit for video virtual try-on. Given a reference person video sequence $\mathbf{I}:=\{ I_{1},...,I_{N}\}\in\mathbb{R}^{3 \times H \times W}$ and a target garment image $G\in\mathbb{R}^{3\times H \times W}$, where $H$ and $W$ denote height and width of the image, and $N$ is the frame length of the sequence, WildVidFit aims to synthesis a realistic video sequence 
$ \Tilde{\mathbf{I}}:=\{\Tilde I_{1},..., \Tilde I_{N} \}\in\mathbb{R}^{3 \times H \times W}$. This video showcases the person wearing the target garment $G$, while maintaining the integrity of all other elements. WildVidFit successfully accomplishes the video try-on task through an image-based approach with two core modules: a one-stage virtual try-on network conditioned on both human motions and garment texture, and a diffusion guidance module for temporal coherence. We start with the preprocessing procedures, followed by a brief introduction on diffusion models. Subsequent subsections further elaborate on the designed one-stage try-on network (Sec.~\ref{sec:one_stage_image_try_on}) and diffusion guidance module (Sec.~\ref{sec:diffusion_guidance}).

\noindent\textbf{Preprocessing of Inputs.}
Drawing inspiration from~\cite{lee2022hrviton, chen2023anydoor}, we propose a method to construct separate representations for the person and the garment, aiming to preserve the individual's identity and accurately reproduce the intricate textures of the garment. Specifically, we obtain the human segmentation map sequence $\mathbf{S}:=\{S_{1},..., S_{N} \}$ and pose maps $ \mathbf{P}:=\{P_{1},..., P_{N} \}$ using off-the-shelf methods~\cite{li2020self,8765346}. Then we produce cloth-agnostic RGB images $ \mathbf{A}:=\{A_{1},..., A_{N} \}$ following the progress described in VITON-HD~\cite{lee2022hrviton}. This process utilizes $\mathbf{P}$ and $ \mathbf{S}$ to effectively remove the original clothing but retains the person identity. Finally the cloth-agnostic RGB images $\mathbf{A}$ and the pose maps $\mathbf{P}$ together form our person representation. For garment representation, in addition to the original garment image $G$, we introduce low-level information represented by edge map $E_g$. $E_g$ is detected by Sobel~\cite{kanopoulos1988design}. We utilized DINO-V2~\cite{oquab2023dinov2} for feature extraction from both garment image $G$ and edge map $E_g$ and then concatenate them into a vector $F_g\in\mathbb{R}^{257\times 2048}$. The dimension of 257 refers to the concatenation of a global token and 256 patch tokens.

\noindent\textbf{Controlled Diffusion Model.}
Diffusion models~\cite{sohl2015deep,ho2020denoising} are a class of generative models that learn the target distribution via an iterative denoising procedure. They consist of a Markovian forward process that progressively corrupts the data sample $\mathbf{x}$ into the Gaussian noise $\mathbf{z}_{T}$, and a learnable reverse process that converts $\mathbf{z}_{T}$ back to $\mathbf{x}$ iteratively. Importantly, diffusion models can be conditioned on various signals like texts or images. A conditional diffusion model $\hat{\mathbf{x}}_{\theta}$ can be trained with a weighted denoising score matching objective:
\begin{equation}
\label{eq:diffusion}
    \mathbb{E}_{\mathbf{x},\mathbf{c},\boldsymbol{\epsilon},t}[{w_t \|\hat{\mathbf{x}}_{\theta}(\alpha_t \mathbf{x} + \sigma_t \boldsymbol{\epsilon}, \mathbf{c}) - \mathbf{x} \|^2_2}],
\end{equation}
where $\mathbf{x}$ is the target data sample, $\mathbf{c}$ is the conditional input, $\boldsymbol{\epsilon} \sim \mathcal{N}(\mathbf{0}, \mathbf{E})$ is the noise term. Here, $\mathbf{E}$ is used to denote the identity matrix. $\alpha_t, \sigma_t, w_t$ are functions of the timestep $t$ according to the formulation of diffusion models. In practice, $\hat{\mathbf{x}}_{\theta}$ is reparameterized as $\hat{\boldsymbol{\epsilon}}_{\theta}$ to predict the noise that corrupts $\mathbf{x}$ into $\mathbf{z}_t :=\alpha_t \mathbf{x} + \sigma_t \boldsymbol{\epsilon}$. During inference, data samples can be generated from Gaussian noise $\mathbf{z}_{T} \sim \mathcal{N}(\mathbf{0}, \mathbf{E})$ using DDIM~\cite{song2020denoising} sampler.

To enable our diffusion model training and inference on limited computational resources without compromising quality and flexibility, we use the pre-trained autoencoder to compress data sample $\mathbf{x}$ into the latent space. 

\subsection{One-stage Image Try-On Network}
\label{sec:one_stage_image_try_on}

The try-on task requires to make controllable image generation where the person wears the target garment while maintaining the original motion. We extend the diffusion model into video try-on task in the form of a conditional image generation task under the joint restriction of the person representation and the garment representation. 

\begin{figure*} [ht]
	\begin{center}
		\includegraphics[width=1.0\linewidth]{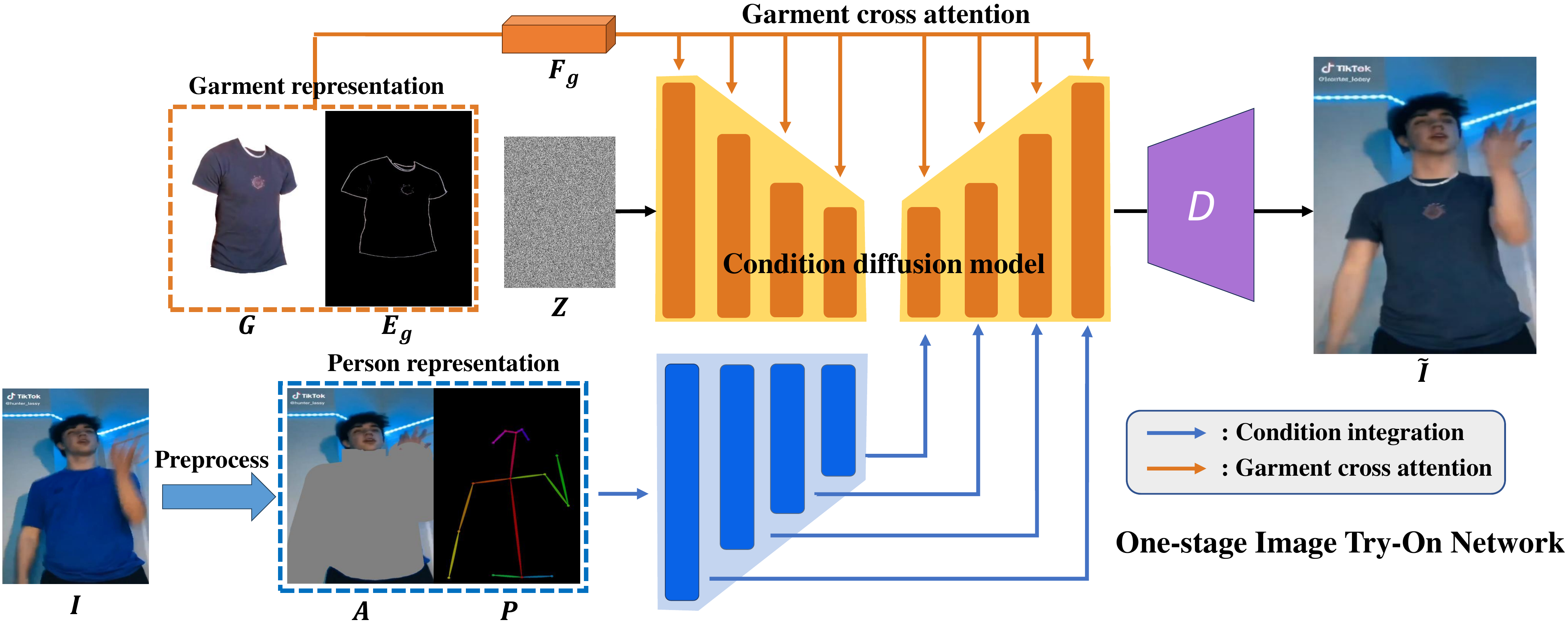} 
	\end{center}
	\caption{\textbf{Overview of the proposed one-stage image try-on network.} First, we extract the person representation and garment representation during preprocessing. The person representation includes the cloth-agnostic image $A$ and the human pose $P$ while the garment representation includes the garment image $G$ and the edge map $E_g$. Then two representations condition the diffusion model in the way of hierarchical fusion in UNet decoder and cross attention respectively.}\label{fig:image_try_on}
\end{figure*}

As shown in Fig.~\ref{fig:image_try_on}, we take the concatenation of the cloth-agnostic image $A$ and the pose image $P$ as the input in condition branch. $A$ and $P$ are critical for preserving the identity of the person as well as the background. The garment representation including the garment image $G$ and its edge map $E_g$, is not aligned with the try-on results. Unlike using the warped garment as the input condition to remove this misalignment ~\cite{morelli2023ladi, gou2023taming}, our network adopts a one-stage paradigm, applying implicit warping via the cross attention between the reference person and the extracted garment feature $F_g$ inspired by TryOnDiffusion~\cite{zhu2023tryondiffusion}. The edge map emphasizes the garment details that need to be maintained. Beneficial in avoiding reliance on the explicit optical flow estimation, our network learns how the garment naturally fits on the person, rather than relying on strict pixel-level transformation, which is essential for generalizing to in-the-wild video images. 


We pick Stable Diffusion~\cite{rombach2022high} as our base architecture but add a condition branch and make cross attention on the garment instead of text. Both the encoder and decoder of the main UNet consist of four blocks with different scale. The architecture of the condition branch is the same as UNet encoder except the first convolution.
We inject the condition signal into the main UNet via convolution.
To preserve the prior knowledge essential for improved generation quality, feature aggregation is only performed in the UNet decoder. Formally, the condition branch $\mathcal{F}$ extracts multi-scale  features $\mathbf{F}_\mathbf{c}=\{\mathbf{F}_\mathbf{c}^1, \mathbf{F}_\mathbf{\mathbf{c}}^2, \mathbf{F}_\mathbf{c}^3, \mathbf{F}_\mathbf{c}^4\}$ from the input condition $\mathbf{c}=\{\mathbf{A},\mathbf{P}\}$, $\mathbf{F}_{\mathbf{c}}$ is corresponding to the output of four blocks. We inject the condition features $\mathbf{F}_\mathbf{c} $ into the decoder feature $\mathbf{F}_{dec}=\{\mathbf{F}_{dec}^1, \mathbf{F}_{dec}^2, \mathbf{F}_{dec}^3, \mathbf{F}_{dec}^4\}$:
\begin{align}
    &\mathbf{F}_\mathbf{c} = \mathcal{F}(\mathbf{A},\mathbf{P}),\\
    &\hat{\mathbf{F}}_{dec}^{i} = Conv(\mathbf{F}_{dec}^{i}, \mathbf{F}_{\mathbf{c}}^{5-i}),\ i\in \{1,2,3,4\},
\end{align}
The objective function in training is the same as Eq (\ref{eq:diffusion}).

\subsection{Temporal Coherent Editing using Diffusion Guidance}
\label{sec:diffusion_guidance}
Generating videos on a frame-by-frame basis will lead to inconsistencies, arising from discrepancies between individual frames. One way to enhance temporal consistency without training a specific temporal module is to leverage the priors in foundational video models. 

\tit{Diffusion Guidance}
One of the diffusion models' notable strengths is their capacity to tailor outputs according to auxiliary information by guiding the sampling process, without fine-tuning the network. Inspired by classifier guidance~\cite{dhariwal2021diffusion, song2020score}, we propose updating the intermediate representation of the sampling process by introducing pre-trained models into the gradients through score functions, thereby achieving coherent video generation. 

As illustrated in Fig.~\ref{fig:guidance}, we introduce the self-supervised video model VideoMAE~\cite{feichtenhofer2022masked} to enhance the coherence of video clips and a self-supervised image model DINO-V2~\cite{oquab2023dinov2} to prevent excessive feature distance between adjacent frames. The video masked autoencoder (VideoMAE), which takes masked videos as input and attempts to reconstruct them by leveraging inter-frame relationships, has been proven to learn strong spatio-temporal representations effectively. This guidance is based on the assumption that smoother videos facilitate easier restoration of masked areas by the autoencoder using information from adjacent frames, resulting in lower reconstruction loss.  We incorporate the score functions into the DDIM~\cite{song2020denoising} process, as detailed in the following formulation:
\begin{equation}
\label{eq:guiding}
\begin{aligned}
    \hat\epsilon_t &= \epsilon_\theta(z_t;t,\mathbf{c}) - w_1\nabla_{z_t} \mathcal{L}_{MAE}(z_t) \\&- w_2\nabla_{z_t} \mathcal{L}_{SIM}(z_t), 
\end{aligned}
\end{equation}
\begin{equation}
    \mathcal{L}_{MAE} = \frac{1}{\Omega} \sum_{p \in \Omega} ||\mathbf{I_t}(p) - \mathbf{\hat{I}_t}(p)||_2, \ \mathbf{I_t}=D(z_t),
\end{equation}
\begin{equation}
    \mathcal{L}_{SIM} = \frac{1}{L-1} \sum_{j=1}^{L-1}dist(f(I^{j+1}_t) - f(I^j_t)),\ \mathbf{I_t}=D(z_t),
\end{equation}
where $z_t\in \mathbb{R}^{L\times H \times W}$ represents video noise at the timestep $t$, $L$ is length of the video clip, $D$ is the decoder in autoencoder. $\mathcal{L}_{MAE}$ is the reconstruction loss of the masked decoded image sequence $\mathbf{\hat{I}_t}$, $p$ denotes the token index and $\Omega$ denotes the set of masked tokens in image sequence $\mathbf{I_t}$. $\mathcal{L}_{SIM}$ is the similarity loss that represents the average distance between two adjacent decoded frames $I^{j+1}_t$ and $I^j_t$ at the timestep $t$. Here $f$ represents the feature extraction function DINO-V2 and we use spherical distance to measure the feature similarity. $w_1$ and $w_2$ are the guidance weights. $\mathcal{L}_{SIM}$ and $\mathcal{L}_{MAE}$ together constitute the temporal loss to guide the iterative denoising procedure. In implementation, we set the masking ratio to 0.7, $w_1=2000$ and $w_2=1000$. Due to the memory limitation, the loss is computed only on the garment area. 

\tit{Long Video Generation}
The length of video clip is fixed in VideoMAE~\cite{feichtenhofer2022masked}. The naive approach to generate long videos is sequential generation, but this approach tends to perform poorly at the junctions of individual clips. In our framework, we adopt a temporal co-denoising strategy to generate longer videos and ensure temporal smoothness. Specifically, we divide the complete reference video into overlapping short video clips, each differing by stride $s$, where $s$ is typically $L//2$ or $L//4$. The co-denoising process can then be represented as follows: at the timestep $t$, the latent $z^j_t$ according the $j^{th}$ frame is the average of all $M$ clips $z_{t,k},k=1,...,M$ including the $j^{th}$ frame:
\begin{equation}
    z^j_t = \frac{1}{M}\sum^M_{k=1}z^j_{t,k},
\end{equation}

\subsection{Other module for Enhanced Performance}
\tit{Autoencoder with Enhanced Mask-Aware Skip Connections} 
The autoencoder directly impacts the image quality. 
In order to preserve fine details better outside the garment region, we fine-tune the autoencoder using the mask-aware skip connection module (EMASC) proposed in \cite{morelli2023ladi}. The EMASC module is defined as follows, taking the garment mask $M$ from the segmentation map $S$:\begin{equation}
D_i = D_{i-1} +  f(E_i) * \lnot m_i, 
\label{eq:emasc}
\end{equation}
where $f$ is a learned non-linear function, $E_i$ is the $i$-th feature map coming from the encoder in autoencoder, $D_i$ is the corresponding $i$-th decoder feature map, and $m_i$ is obtained by resizing the mask $M$ to adapt the spatial dimension. Here, $\lnot m_i$ yields the logical negation of $m_i$, i.e., obtaining the unmasked region.

\tit{Fully Cross-frame Attention}
We replace self-attention by fully cross-frame attention in the UNet while making sequential inference to increase the spatial-temporal coherency as proposed in~\cite{zhang2023controlvideo}.
{
\begin{equation}
\begin{aligned}
    \text{Attention}(Q,K,V)=\text{softmax}(\frac{QK^T}{\sqrt{d}})V,\\
    \text{ where }Q=W^Q z_t, \  K=W^K z_t,\  V=W^V z_t,
    \label{eq:full_attn}
\end{aligned}
\end{equation}
}
Here $z_t = \{z_t^i\}_{i=1}^{L}$ denotes all $L$ latent frames of the video clip at the timestep $t$, while $W^Q$, $W^K$, and $W^V$ project $z_t$ into query, key, and value, respectively.

\section{Experiments}
Our experiments are divided into four parts. Firstly, we demonstrate the superiority of our one-stage virtual try-on network through image-based virtual try-on experiments in Sec.~\ref{sec:image_try_on}. Secondly, we validate the efficacy of our image-based approach on the public VVT dataset~\cite{dong2019fw} in Sec.~\ref{sec:vtt_try_on}.
Thirdly, to showcase the robustness and generalizability of the proposed WildVidFit framework, we conduct video virtual try-on on the TikTok dataset~\cite{jafarian2022self}, which is introduced in Sec.~\ref{sec:wild_video}. Finally, ablation studies are conducted in Sec.~\ref{sec:ablation_study}. 
\subsection{Experiment Setup}
\tit{Datasets} Our image virtual try-on experiments are conducted on two existing high-resolution virtual try-on benchmarks VITON-HD~\cite{choi2021viton} and DressCode~\cite{morelli2022dress}. VITON-HD contains 13679 garment-person pairs, 11647 for training while remaining 2023 for testing. For DressCode dataset, we use the upper subset of it and 15365 image pairs are split into 13564/1801 training/testing pairs. 

In the video try-on task, we evaluate our WildVidFit framework on the VVT~\cite{dong2019fw} and TikTok dataset~\cite{jafarian2022self}. The VVT dataset~\cite{dong2019fw} includes 791 videos, each with a resolution of $256\times 192$. It is divided into a training set with 159,170 frames and a test set with 30,931 frames. However, the VVT dataset~\cite{dong2019fw} primarily features monotonous and simple human poses against predominantly white backgrounds. In contrast, the TikTok dataset~\cite{jafarian2022self} comprises over 300 dance videos that captures a single person performing complex dance moves with intricate limb occlusions and dynamic postures.  We selected 165 videos with clear upper body views from this collection. Garment-person pairs are created from TikTok frames using Grounded-SAM~\cite{kirillov2023segany}. The training set includes 130 videos and 34,933 frames, while the test set contains 35 videos and 9816 frames.

\tit{Training and Testing} The main UNet of our one-stage try-on network inherits the parameter of Stable Diffusion~\cite{rombach2022high}. 
When adapting the network for the try-on task, we train only the UNet decoder and the condition branch $\mathcal{F}$, while keeping the encoder frozen. This training strategy preserves the priors and avoids overfitting. We train the network for 100K iterations with a batch size of 16 using AdamW optimizer~\cite{loshchilov2017decoupled}. 
The learning rate is set to $5e^{-5}$. The resolution is $512\times 384$ on the VITON-HD, DressCode and TikTok Dataset, while the VVT dataset maintains its original resolution of $256\times 192$.

At inference time, We use DDIM~\cite{song2020denoising} as the sample method, and the total steps is 30. Classifier-free guidance sample~\cite{ho2022classifier} is able to strength the influence of conditions on the generated images. We use the conditional classifier-free guidance on the garment feature $F_g$, the guidance sacle is set to 2. 

\subsection{Image Try-on Results}
\label{sec:image_try_on}
\tit{Evaluation Metrics} Following previous studies~\cite{lee2022hrviton}, we make quantitative evaluation on both paired and unpaired setting. In the paired setting, we employ SSIM~\cite{wang2004image} and LPIPS\cite{zhang2018unreasonable} as evaluation metrices. In the unpaired setting, where ground truth is unavailable, we evaluate realism using the Fréchet Inception Distance (FID)~\cite{heusel2017gans}, Kernel Inception Distance (KID)~\cite{binkowski2018demystifying} scores and user study. For user study, 100 samples are randomly selected and 50 volunteers are asked to select the one of best quality among different methods.


\begin{figure}[ht]
  \centering
  \includegraphics[width=0.9\textwidth]{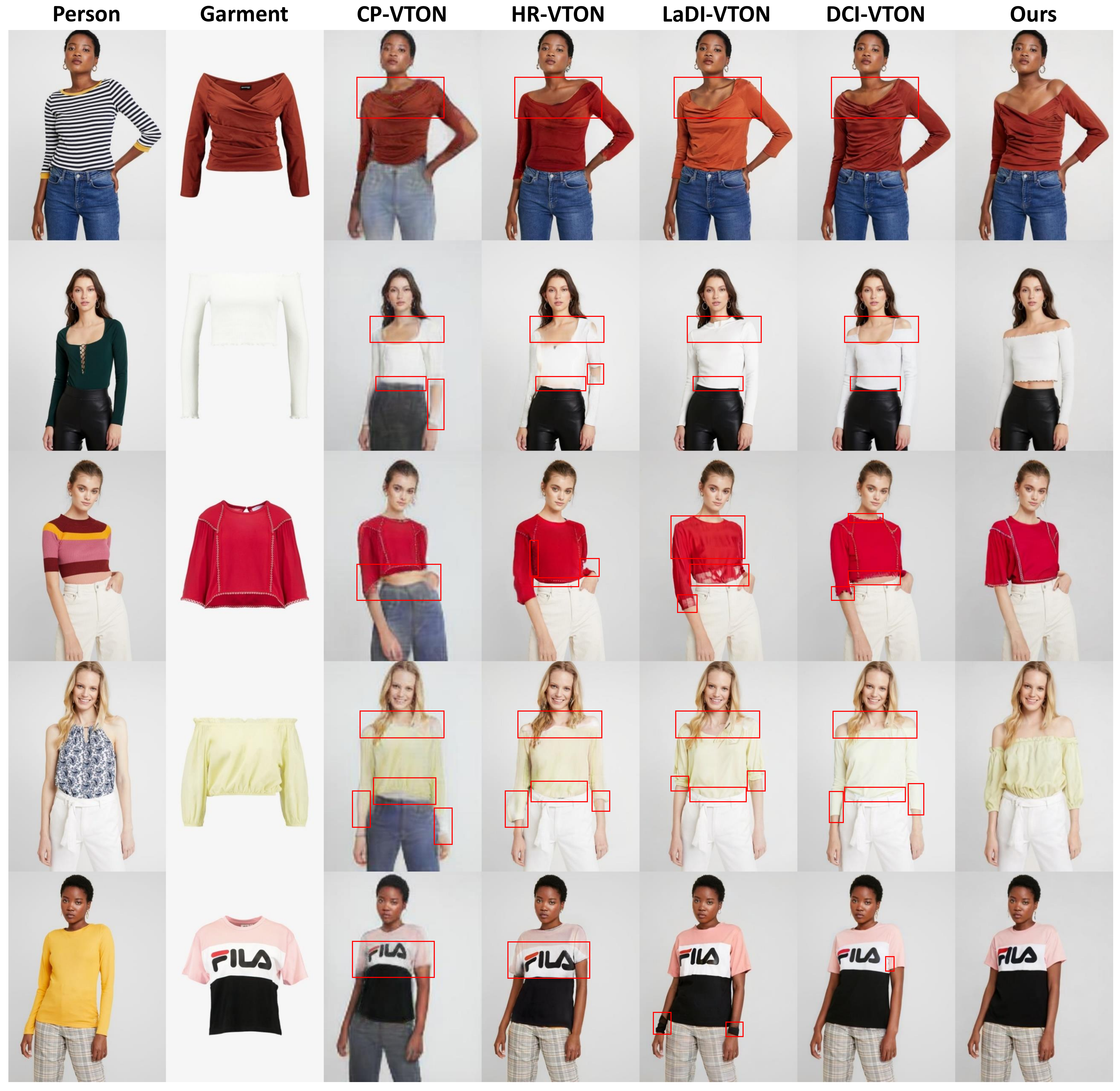}
  \caption{\textbf{Qualitative comparison on VITON-HD dataset.} Zoom in for best view.}
  \label{fig:image_comparison}
\end{figure}

\begin{table}[h]
    \caption{Quantitative comparison with baselines on VITON-HD dataset.}
    \centering
    \setlength{\tabcolsep}{2mm}{
    \begin{tabular}{l|ccccc}
    \toprule
    Methods & SSIM↑\ & LPIPS↓ & FID↓ & KID↓ & User↑\\ 
    \midrule
    CP-VTON~\cite{wang2018toward} & 0.785 & 0.2871 & 48.86 & 4.42 & $3.86\%$ \\
    HR-VTON~\cite{lee2022hrviton} & 0.878 & 0.0987 & 11.80 & 0.37 & $6.62\%$ \\
    Ladi-VTON~\cite{morelli2023ladi} & 0.871 & 0.0941 & 13.01 & 0.66 & $16.02\%$ \\
    DCI-VTON~\cite{gou2023taming} & 0.882 & 0.0786 & 11.91 & 0.51 & $12.18\%$ \\
    WildVidFit(Ours)& \textbf{0.883} & \textbf{0.0773}  & \textbf{8.67}  & \textbf{0.10} & $\textbf{61.32\%}$\\
    \bottomrule
    \end{tabular}}
    \label{table:viton_compare}
\end{table}

\begin{table}[h]
    \caption{Quantitative comparison with baselines on DressCode-Upper dataset.}
    \centering
    \setlength{\tabcolsep}{2mm}{
    \begin{tabular}{l|ccccc}
    \toprule
    Methods & SSIM↑ & LPIPS↓ & FID↓ & KID↓ & User↑\\ 
    \midrule
    CP-VTON~\cite{wang2018toward} & 0.820 & 0.2764 & 57.70 & 4.56 & $0.00\%$ \\
    HR-VTON~\cite{lee2022hrviton} & 0.924 & 0.0605 & 13.80 & 0.28 & $5.16\%$ \\
    Ladi-VTON~\cite{morelli2023ladi} & 0.915 & 0.0620 & 16.71 & 0.61 & $26.20\%$ \\
    WildVidFit(Ours)& \textbf{0.928} & \textbf{0.0432}  & \textbf{12.48}  & \textbf{0.19} & $\textbf{68.64\%}$\\
    \bottomrule
    \end{tabular}}
    \label{table:dresscode_compare}
\end{table}

\tit{Comparison with State-of-the-Art Models} We compare our method against CP-VTON~\cite{wang2018toward}, HR-VITON~\cite{lee2022hrviton}, LaDI-VTON~\cite{lee2022hrviton} and DCI-VTON~\cite{gou2023taming} using their official codes and checkpoints. Since no available checkpoint or code for DCI-VTON on DressCode dataset, we skip this comparison.

Qualitative comparison on VITON-HD~\cite{choi2021viton} is exhibited in Fig.~\ref{fig:image_comparison}. Our method consistently preserves essential clothing characteristics, setting it apart from other methods that often falter with inadequate feature retention and evident blurring, as illustrated  in rows 2 and 3. 
Notably, for garments with intricate folds, our model adeptly retains the complex textures, while competing methods tend to produce overly smoothed results, as observed in rows 1 and 4. Such distinctions underscore our model's capacity to discern the nuanced interplay between human and garment. Examples on DressCode are presented in the Appendix.

Table ~\ref{table:viton_compare} and Table ~\ref{table:dresscode_compare} show quantitative comparison with previous methods, confirming the superiority of our method in image visual quality under both paired and unpaired evaluation. This reveals that our method achieves state-of-the-art performance in the image-level virtual try-on task.

\subsection{Video Try-On Results on VVT Dataset}
~\label{sec:vtt_try_on}

\tit{Evaluation Metrics}  We use Video Frechet Inception Distance (VFID) to measure the generation quality and temporal consistency following~\cite{dong2019fw}. VFID is a variant of FID, extracting feature vector of video clips for metric computation by pre-trained video backbone I3D~\cite{I3D}. Each video clip includes 36 frames. Also we adds a user survey for subjective evaluation, with settings consistent with image try-on evaluation above.

\tit{Comparison with State-of-the-Art Models} We compare our method with video-based method ClothFormer~\cite{han2019clothflow} and imaged-based methods HR-VTON~\cite{lee2022hrviton} and LaDI-VTON~\cite{morelli2023ladi}. 
The quantitative experiment, as shown in Table ~\ref{table:video_comparison}, demonstrates that our method outpaces image-based approaches and matches the performance of the video-based ClothFormer. This underscores the robustness of our one-stage image virtual try-on network and the effectiveness of diffusion guidance in maintaining temporal consistency. The visual comparison can be seen in the Appendix. 


\begin{table}[h]
    \caption{Quantitative comparison on the VVT and TikTok dataset.}
    \centering
    \begin{tabular}{l|ccc}
    \toprule
    Methods & Dataset & VFID↓ & User↑ \\ 
    \midrule
    HR-VTON~\cite{lee2022hrviton} & VVT & 4.852   &  9.46\%    \\
    LaDI-VTON~\cite{morelli2023ladi} & VVT & 4.442    & 4.24\%    \\
    ClothFormer~\cite{jiang2022clothformer} & VVT & \textbf{4.192}    & \textbf{46.44\%}   \\
    WildVidFit(Ours) & VVT & 4.202 &   39.86\% \\
    \midrule
    HR-VTON~\cite{lee2022hrviton} & TikTok & 25.43 & 0.00\%\\
    LaDI-VTON~\cite{morelli2023ladi} & TikTok & 14.24 & 26.90\%\\
    WildVidFit(Ours) & TikTok & \textbf{9.87} & \textbf{73.10\%}\\
    \bottomrule
    \end{tabular}
    \label{table:video_comparison}
\end{table}

\begin{figure}[h]
  \centering
   \includegraphics[width=0.9\linewidth]{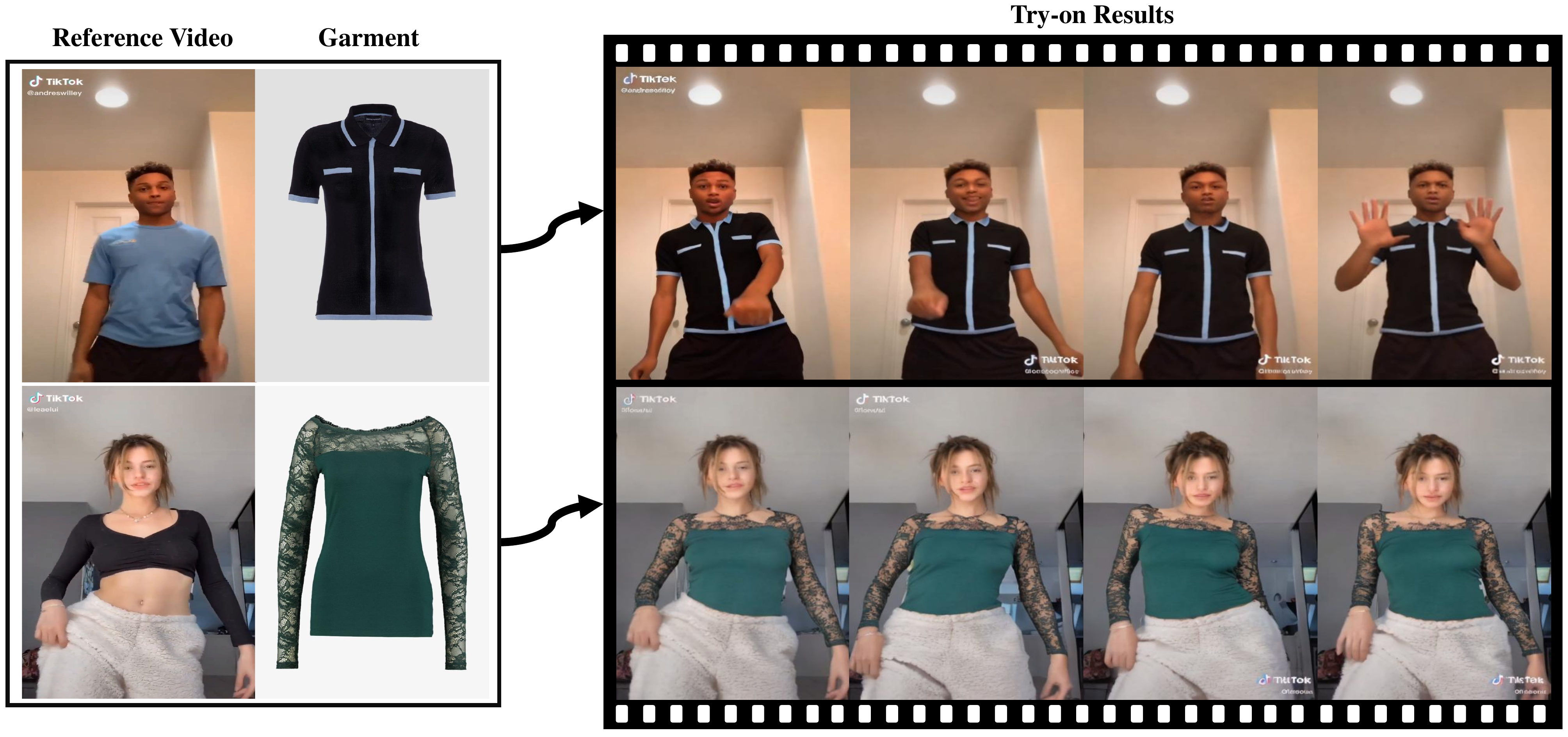}
   \caption{\textbf{Cross-dataset video try-on results}, given a reference video from TikTok dataset and a garment item from DressCode (1st row) and VITON-HD (2nd row) dataset. Zoom in for optimal viewing.}
   \label{fig:cross_dataset}
\end{figure}

\begin{figure*} [h]
	\begin{center}
		\includegraphics[width=1.0\linewidth]{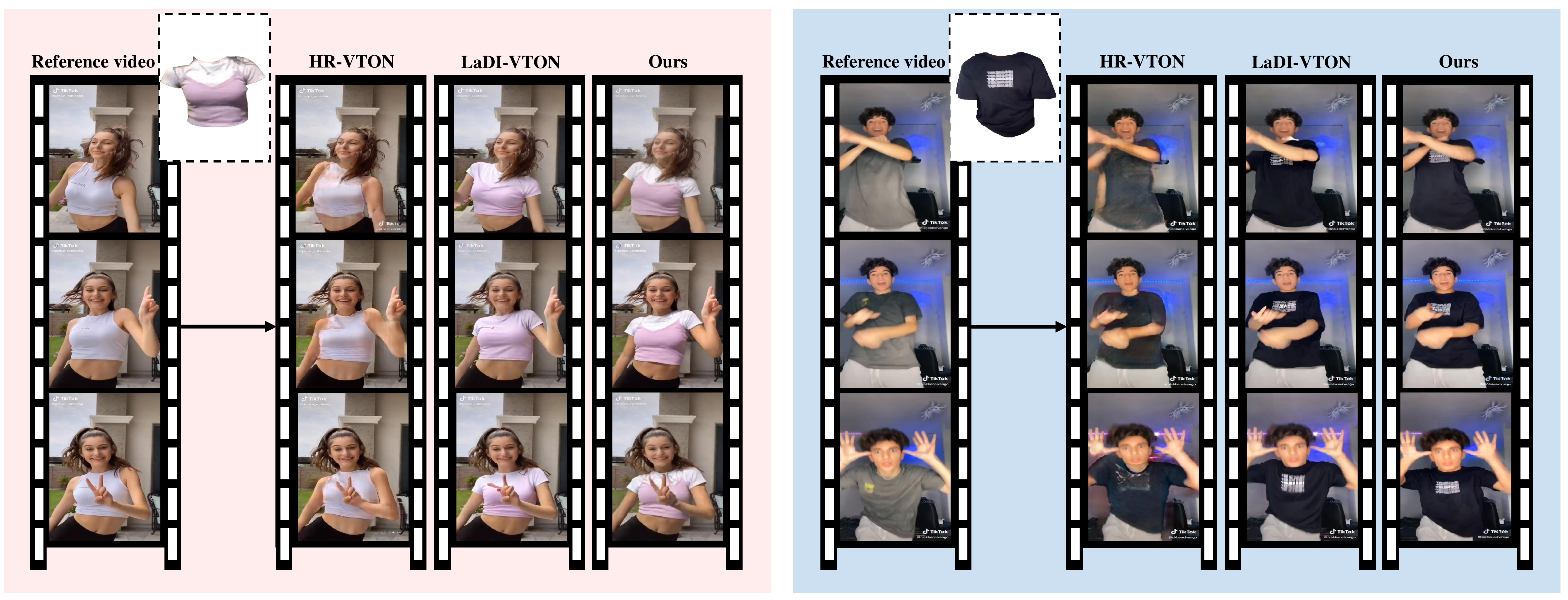} 
	\end{center}
        \caption{\textbf{Qualitative comparison on the TikTok dataset.} Our approach can reproduce the details of clothing under dance movements, while other methods perform poorly in cases of limb occlusion. Zoom in for optimal viewing.}\label{fig:tiktok_compare}
\end{figure*}

\begin{figure}[h]
  \centering
   \includegraphics[width=0.9\linewidth]{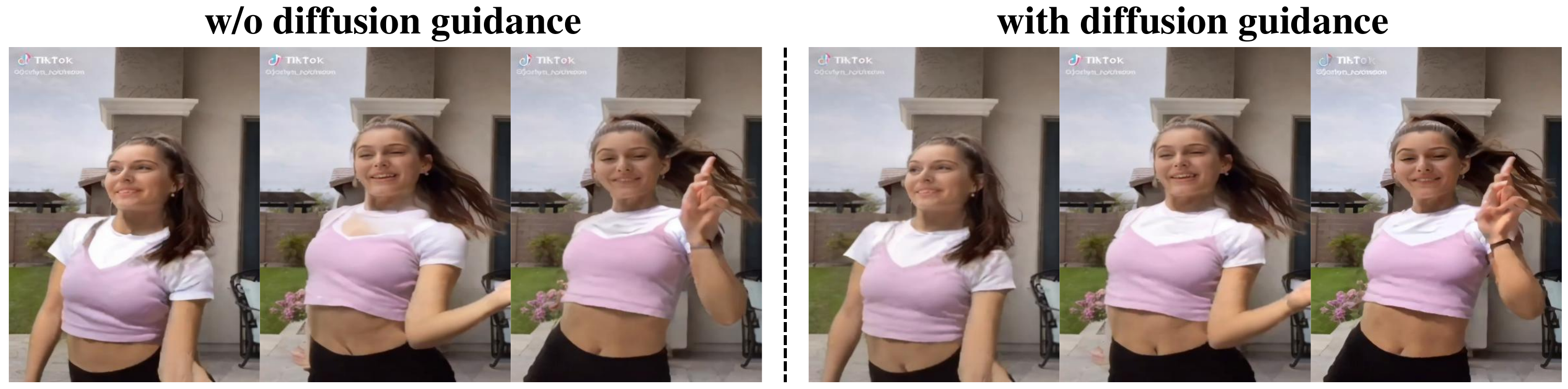}
   \caption{\textbf{Effects of diffusion guidance.} The guidance module enhances the smoothness and mitigates artifacts on the garment by incorporating overall information.}
   \label{fig:ablation}
\end{figure}

\subsection{In-the-Wild Video Virtual Try-On}
\label{sec:wild_video}
\tit{Joint Training on Multiple DataSets}
Dance videos from TikTok dataset~\cite{jafarian2022self} can effectively evaluate the capability of our method in handling wild videos. To enhance the model's generalization ability for processing the TikTok videos, we conduct joint training using three datasets: VITON-HD~\cite{choi2021viton}, DressCode~\cite{morelli2022dress}, and TikTok\cite{jafarian2022self}. Benefit from this, we are able to transfer the garments from VITON-HD and DressCode onto the TikTok videos as shown in Fig.~\ref{fig:cross_dataset}. This to some extent demonstrates the robustness of our method.

\tit{Comparison with State-of-the-Art Models} Since there is no public source or commercial software for video try-on, we are compelled to compare our method with image-based methods, i.e, HR-VTON~\cite{lee2022hrviton} and LaDI-VTON~\cite{morelli2023ladi}. For fair comparison, HR-VTON and LaDI-VTON also adopt the strategy of joint training on the three datasets mentioned above.
Fig.~\ref{fig:tiktok_compare} visualizes the comparison results.
It can be observed that the GAN-based method HR-VTON completely fails, whereas LaDI-VTON, despite leveraging the foundational capabilities from Stable Diffusion~\cite{rombach2022high}, performs poorly in cases of limb occlusion. This is primarily due to the challenges in warping. Our method, on the other hand, accurately reproduces the details of the garments, ensuring that the garment fits well with the person's motions. Table ~\ref{table:video_comparison} also shows the clear superiority of our method. 

\begin{table}[h]
\caption{Ablation study for edge maps and CFG on VITON-HD dataset.}
    \centering
    \setlength\tabcolsep{2mm}{
    \begin{tabular}{cc|cc}
    \toprule
    Edge maps\quad \ & Guidance scale\quad &FID↓ & KID↓ \\ 
    \midrule
    \xmark & 2 & 8.93 & 0.12 \\
    \cmark & 1 & 9.47 & 0.17 \\
    \cmark & 2 & \textbf{8.67} & \textbf{0.10} \\
    \cmark & 3 & 8.68 & 0.10 \\
    \bottomrule
    \end{tabular}}
    \label{table:guidance_scale}
\end{table}

\begin{table}[h]
\caption{Ablation study for temporal modules on TikTok dataset.}
\setlength\tabcolsep{2pt}%
    \centering
    \begin{tabular}{l|c}
    \toprule
    Methods & VFID↓ \\
    \midrule
    Image-based & 13.45  \\
    +~Fully cross-frame attention & 12.14 \\
    +~Guidance with $\mathcal{L}_{MAE}$ & 10.64\\
    +~Guidance with $\mathcal{L}_{MAE}$ and $\mathcal{L}_{SIM}$& 10.28 \\
    +~Temporal co-denoising strategy& \textbf{9.87}\\
    \bottomrule
    \end{tabular}
    \label{table:ablation}
\end{table}


\subsection{Ablation Study}
\label{sec:ablation_study}
In this section, we analyze the effectiveness the edge map as well as the classifier-free guidance scale (CFG) and the contribution of each module to the temporal consistency. 

\tit{Effectiveness of Edge Maps and Guidance Scale} In the virtual try-on task, we aim to enhance the preservation of garment textures. We conducted an ablation experiment on the guidance scale of garment feature $F_g$ and the effectiveness of the edge map $E_g$. As shown in Table~\ref{table:guidance_scale}, the introduction of edge maps using cross-attention has resulted in improvement. And our method achieved the best results when the guidance scale was set to 2, yielding a KID score of 8.67 and a FID score of 0.10.

\tit{Effectiveness of Temporal Module} We conducted an ablation study to analyze the designed guided diffusion module and other temporal techniques. The baseline is the direct prediction of image sequences, and then we sequentially incorporate fully cross-frame attention, guidance with $\mathcal{L}_{MAE}$, guidance with $\mathcal{L}_{SIM}$ and co-denoising strategy, in order to analyze the effectiveness of each module. It can be seen that each module can bring improvement. 
Fig.~\ref{fig:ablation} shows that the images generated without diffusion guidance exhibit flaws in the garment. This supports the idea that the propose diffusion guidance module can not only enhance the smoothness of videos, but also use the overall video information to rectify some inconsistencies in single images.


\section{Conclusions}
To effectively tackle the complexities of video virtual try-on in the wild , we introduce WildVidFit, an innovative image-based virtual try-on framework. Specifically designed to manage the challenged poses by frequent movement and significant occlusions common in wild video footage, WildVidFit employs a one-stage, detail-oriented image diffusion model conditioned on both garment and person. The training with a large number of image pairs endows our model with robust performance. Moreover, WildVidFit achieves inter-frame consistency through the technique of diffusion guidance, thereby enabling successful video try-on within a predominantly image-based framework. Our comprehensive experiments reveal that our method not only achieves state-of-the-art performance in image try-on task but also marks a significant foray into video try-on in the wild.

\section*{Acknowledgements}
This work was supported in part by the National Natural Science Foundation of China (NO.~62322608, NO.~62325605), in part by the Fundamental Research Funds for the Central Universities under Grant 22lgqb25, in part by the CAAI-MindSpore Open Fund, developed on OpenI Community, and in part by the Open Project Program of State Key Laboratory of Virtual Reality Technology and Systems, Beihang University (No.VRLAB2023A01). 

\clearpage  

%
%
\bibliographystyle{splncs04}
\bibliography{arxiv}

\newpage
\appendix
\begin{center}
      {\bf APPENDIX}
    \end{center}

In this document, we provide additional materials to supplement our main text. In Appendix A, we show more visualization results, including a qualitative comparison of image try-on on DressCode dataset~\cite{morelli2022dress}, a qualitative comparison of video try-on on VVT dataset~\cite{dong2019fw} and additional video try-on results on TikTok dataset~\cite{jafarian2022self} generated by our WildVidFit framework. Then we show failure cases and discuss the limitations of our method in Appendix B.

\section{More Visualization Results}
\subsection{Image Try-on Results on DressCode}
\begin{figure*} [ht]
	\begin{center}
		\includegraphics[width=1.0\linewidth]{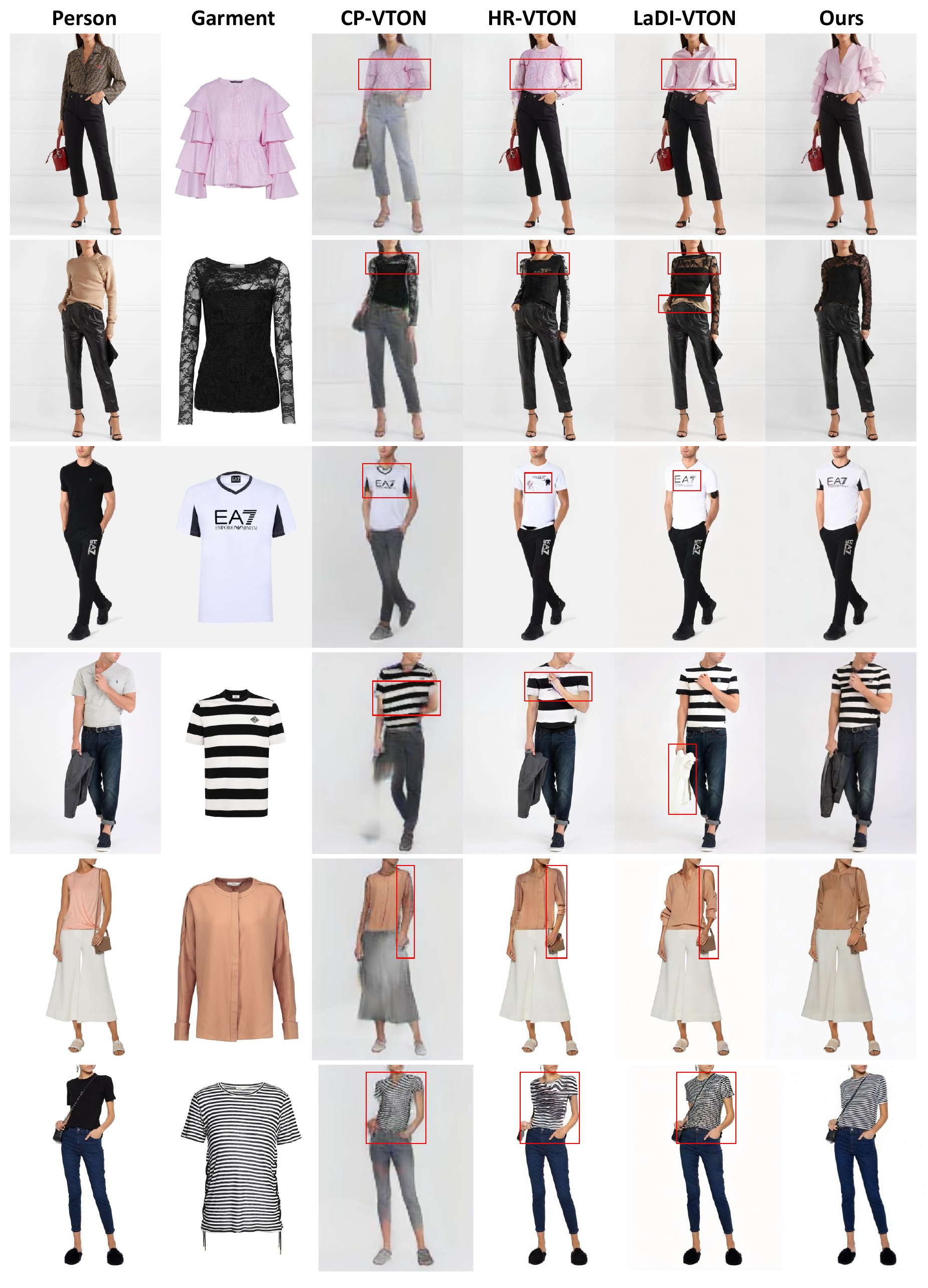} 
	\end{center}
        \caption{\textbf{Qualitative comparison on DressCode dataset.} Please zoom in for best view.}
        \label{fig:dresscode_compare}
\end{figure*}

Fig.~\ref{fig:dresscode_compare} provides qualitative results to demonstrate the superiority of our one-stage try-on network. Our network can effectively handle the garments with complex designs, accurately reproducing their features for impressive try-on results. This includes, for example, the multi-layered wrinkled sleeves in row 1 and the lace garment in row 2. Such designs are challenging to restore through warping, hence other methods fail to make reasonable predictions. Moreover, on striped garments, as in rows 3 and 4, other methods result in blurred outcomes, while our approach successfully maintains the textures. As demonstrated in Tab.~\ref{table:image_comparison}, the performance of our method significantly surpasses that of others.

\begin{figure*} [h]
    \vspace{-11pt}
	\begin{center}
		\includegraphics[width=1.0\linewidth]{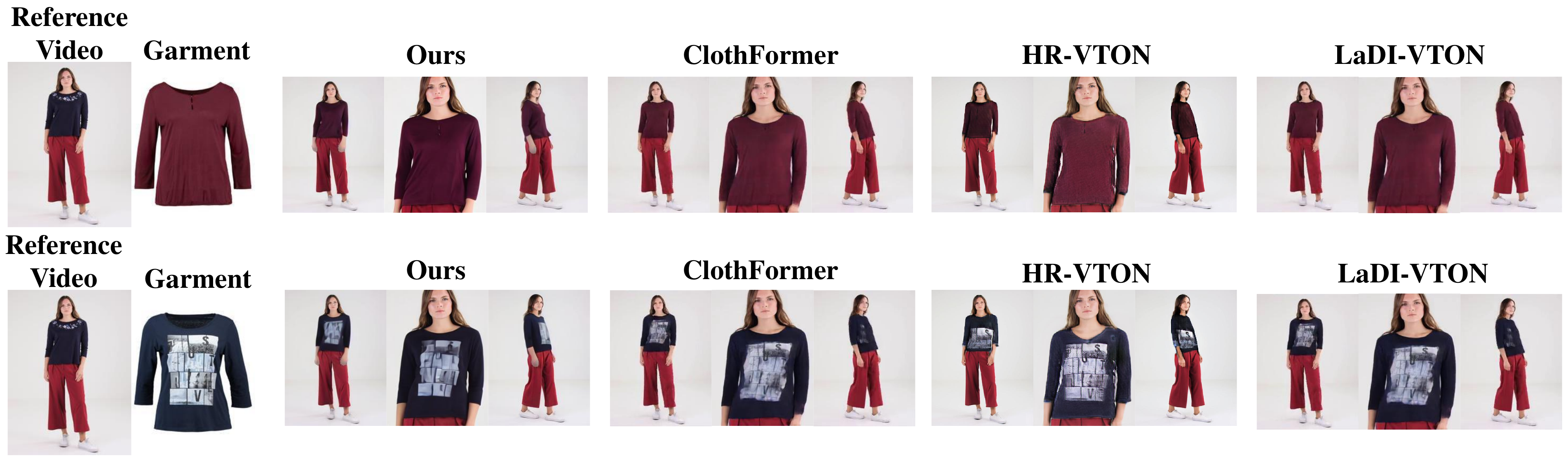} 
	\end{center}
        \vspace{-11pt}
	\caption{\textbf{Qualitative comparison with state-of-the-art methods ClothFormer~\cite{jiang2022clothformer}, HR-VOTN~\cite{lee2022hrviton} and LaDI-VTON~\cite{morelli2023ladi} on the VVT dataset.} Our model achieved robust results under various poses, thereby forming a coherent video sequence.}\label{fig:vtt}
 \vspace{-11pt}
\end{figure*}

\subsection{Qualitative Comparison on VVT Dataset}
In the video try-on task, we compare our method on VVT dataset~\cite{dong2019fw} against HR-VTON~\cite{lee2022hrviton}, LaDI-VTON~\cite{morelli2023ladi} and ClothFormer~\cite{jiang2022clothformer}. As shown in Fig.~\ref{fig:vtt}, although our method has some difficulties with very fine details like small text on clothes due to implicit warping, it outperforms others in robustness when generating with various poses. In comparison, HR-VTON and ClothFormer tend to produce blurred textures in side views. In terms of temporal consistency, HR-VTON and LaDI-VTON naturally lag behind due to their lack of a temporal module, whereas our method performs on par with ClothFormer.

\subsection{More Video Try-on Results on TikTok Dataset}
\begin{figure*} [ht]
	\begin{center}
		\includegraphics[width=1.0\linewidth]{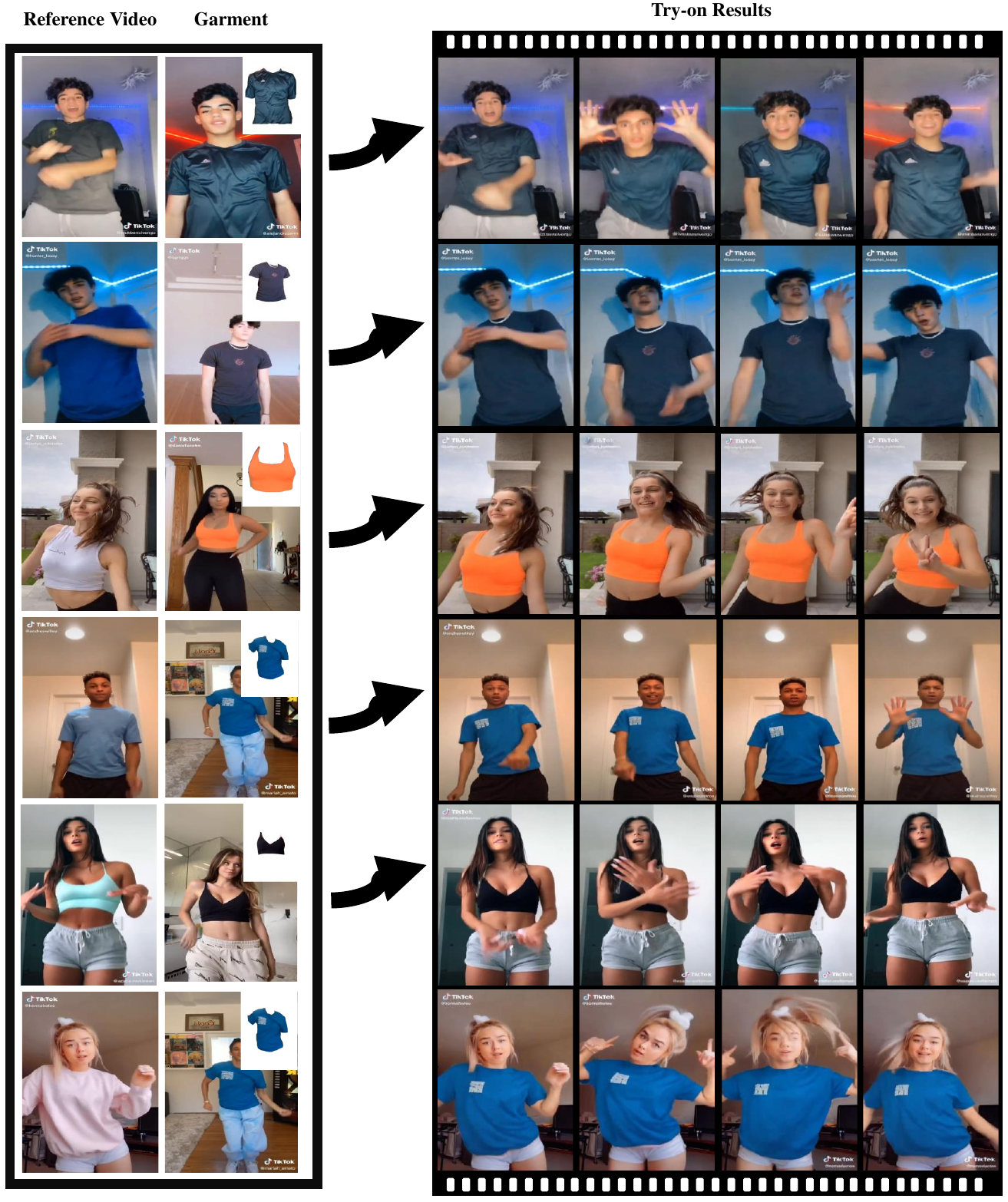} 
	\end{center}
        \caption{\textbf{More examples of our virtual try-on results on real-life TikTok videos.} For optimal viewing, please zoom in or see the supplementary videos.}
        \label{fig:tiktok_more}
\end{figure*}

Fig.~\ref{fig:tiktok_more} showcases more of our video virtual try-on results on the TikTok dataset~\cite{jafarian2022self}. Our method maintains a good appearance of the garments and is able to produce continuous, smooth videos. (\textbf{Notes}: Recommend to see the supplementary videos.) These examples further illustrate the effectiveness of our method in processing wild videos.

\begin{figure} [h]
    \vspace{-10pt}
	\begin{center}
		\includegraphics[width=0.95\linewidth]{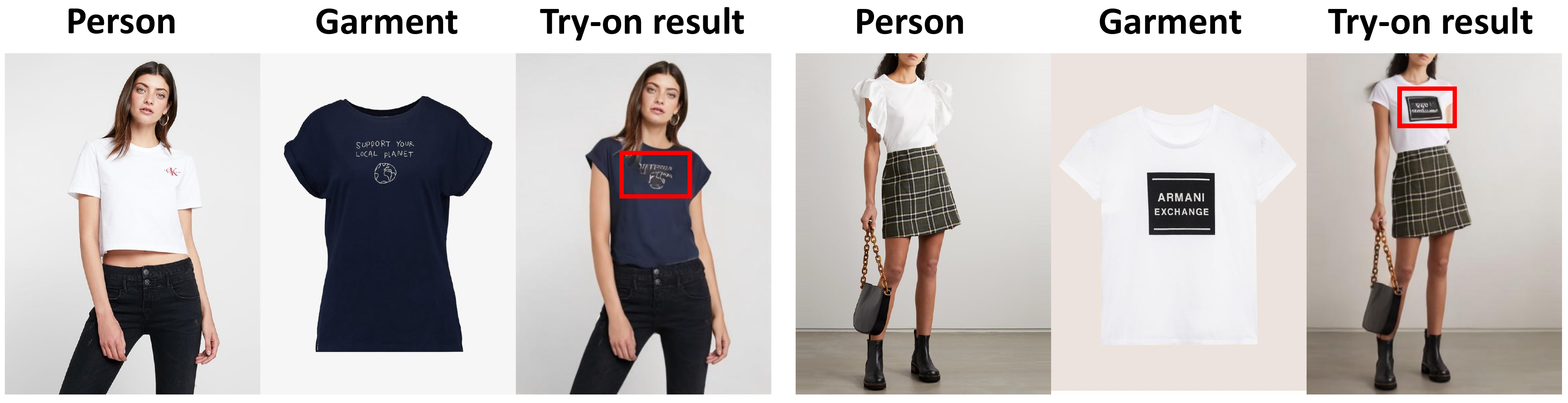} 
	\end{center}
        \vspace{-5pt}
        \caption{\textbf{Failure cases of image try-on results} on DressCode (1st row) and VITON-HD (2nd row). Please zoom in to see differences highlighted by red boxes.}
        \label{fig:failure_case_0}
        \vspace{-15pt}
\end{figure}

\begin{figure} [h]
	\begin{center}
		\includegraphics[width=.9\linewidth]{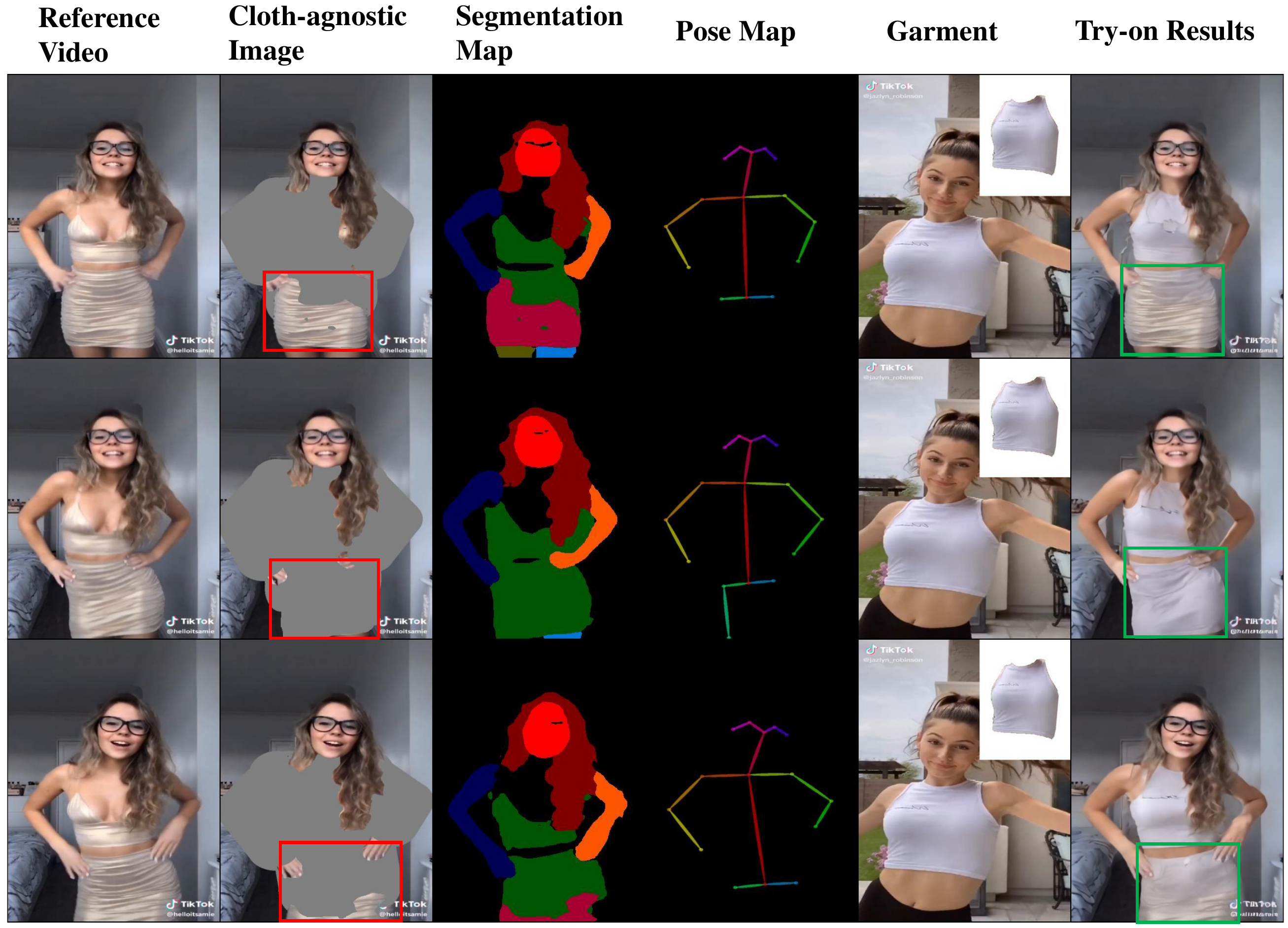} 
	\end{center}
        \vspace{-5pt}
        \caption{\textbf{Failure cases of inaccurate segmentation results.} The inappropriate masked regions are highlighted by red boxes and the imperfect restoration of the lower garment caused by this is indicated by green boxes. For optimal viewing, please zoom in or see the supplementary videos.}
        \label{fig:failure_case_seg}
        \vspace{-13pt}
\end{figure}

\section{Failure Cases and Limitations}
Despite some satisfying results, our method exhibits certain limitations. As illustrated in Fig.~\ref{fig:failure_case_0}, our one-stage try-on network fails to reproduce tiny and complex patterns, such as little writing on the garment. This might stem from the implicit warping processed in the latent space, where overly detailed textures are lost during compression. Another limitation arises from inaccurate parsing results. As shown in Fig.~\ref{fig:failure_case_seg}, the segmentation algorithm fails to distinguish between upper and lower garment, resulting in unreasonable cloth-agnostic region where the lower garment is inappropriately masked (highlighted by red boxes). Such erroneous and discontinuous segmentation maps not only lead to imperfect restoration of the lower garment (highlighted by green boxes) but also adversely affect the virtual try-on result on the upper garment. For a clearer understanding, please refer to the supplementary video.

\end{document}